\documentclass[10pt,twocolumn,letterpaper]{article}

\usepackage[pagenumbers]{cvpr} 

\usepackage{url}
\usepackage{multirow}
\usepackage{xspace}
\usepackage{graphicx}
\usepackage{microtype}
\usepackage{amsmath}
\usepackage{amssymb}
\usepackage{mathtools}
\usepackage{amsthm}
\usepackage{booktabs}
\usepackage{color}
\usepackage{colortbl}
\definecolor{Gray}{gray}{0.95}

\usepackage[pagebackref=true,colorlinks,citecolor=brown]{hyperref}

\usepackage[mathscr]{euscript}

\def\matR{{\mathbb{R}}}

\def\OURS{{OpenREAD}\xspace}

\title{OpenREAD: Reinforced Open-Ended Reasoning for End-to-End Autonomous Driving with LLM-as-Critic}

\author{%
Songyan Zhang$^{1}$\thanks{Equal Contribution}, ~ Wenhui Huang$ ^{2*}$, ~ Zhan Chen$^1 $, \\
~ Chua Jiahao Collister$^1 $, ~ Qihang Huang$^1 $, ~ Chen Lv$ ^{1}\thanks{Corresponding author}$
\\[0.24cm]
$ ^1 $ Nanyang Technological University, Singapore
~~~
$ ^2 $ Harvard University, USA
}
\vspace{-10pt}
\begin{document}

\twocolumn[
\maketitle
\vspace{-20pt}
\begin{center}
\url{https://github.com/wyddmw/OpenREAD}
\end{center}
\vspace{5pt}
]

\begin{abstract}
Recently, two-stage fine-tuning strategies, e.g., acquiring essential driving knowledge through supervised fine-tuning (SFT) and further enhancing decision-making and planning via reinforcement fine-tuning (RFT), have shown strong potential in advancing the knowledge-driven autonomous driving (AD) paradigm. However, the learning nature of SFT still limits the generalization of reasoning, thereby constraining the full potential of driving performance. Meanwhile, current RFT approaches are primarily applied to downstream tasks, since scene understanding is an open-ended problem where corresponding rewards are difficult to quantify. To address these limitations, we propose \OURS, an \textbf{OPEN}-ended \textbf{RE}asoning reinforced vision-language model (VLM)-based autonomous driving (\textbf{AD}) framework that enables end-to-end RFT across the full spectrum from high-level reasoning to low-level trajectory planning. Specifically, we begin by constructing large-scale Chain-of-Thought (CoT) annotations on open-source driving-related knowledge datasets, and employ the powerful Qwen3 large language model (LLM) as the critic in RFT to quantify reasoning quality for open-ended questions during reward modeling. Extensive experiments confirm that joint end-to-end RFT yields substantial improvements in both upstream and downstream tasks, enabling \OURS to achieve state-of-the-art performance on reasoning and planning benchmarks. 



\end{abstract}    

\begin{figure}[ht]
\centering
\includegraphics[width=0.48\textwidth]{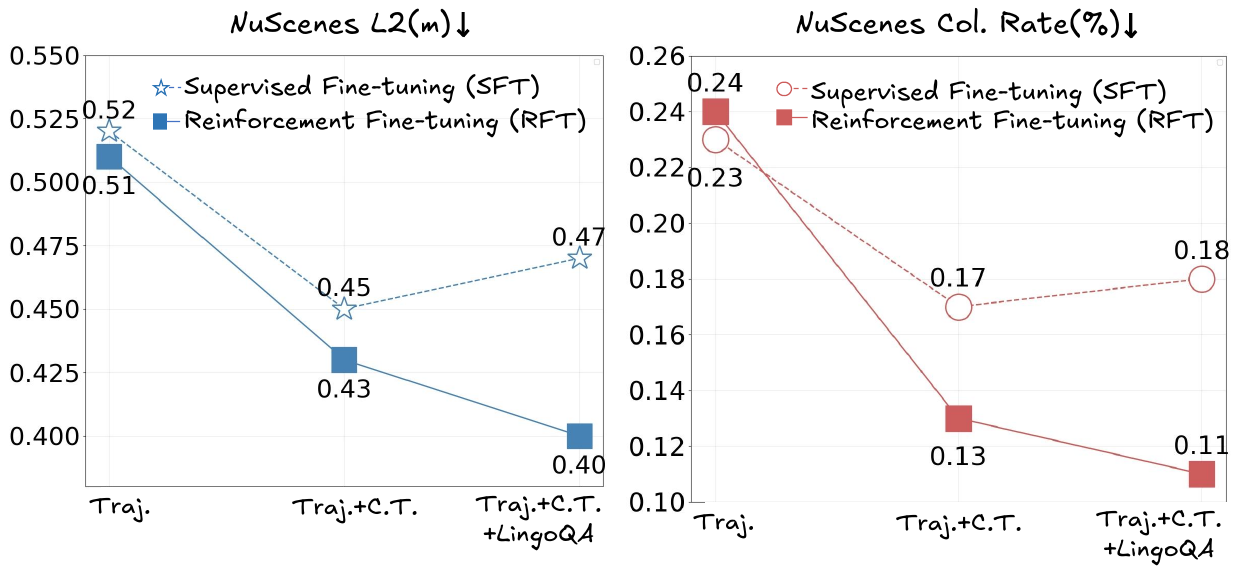}
\caption{\textbf{Comparison between RFT and SFT with increased driving-related knowledge}. Traj. denotes the data of trajectory planning only, and C.T. denotes the data of counterfactual trajectory analysis. Extending driving knowledge through RFT leads to a notable improvement on the NuScenes dataset compared with SFT.}
\label{fig:rft_curve}
\end{figure}

\section{Introduction}
Autonomous driving has witnessed remarkable progress in recent years, largely propelled by breakthroughs in machine learning algorithms such as convolutional neural networks~\cite{resnet, convnext}, and transformers~\cite{transformers, swin-transformers}. This progress has been further promoted by the rapid growth of large-scale, high-quality datasets~\cite{carla, nuscenes, kitti, waymo}, tailored for the development of individual modules—including perception, prediction, and planning, \etc. More recently, end-to-end (E2E) data-driven paradigms have been proposed to integrate these traditionally independent components into a unified framework~\cite{vad, uniad, vadv2, paradrive, bev_planner}. Such E2E data-driven approaches enable the joint optimization across perception, prediction, and planning modules, facilitating scalability in both model parameters and training data, and obtaining substantial performance improvements. However, as discussed in \cite{ad_knowledge}, contemporary methods still face significant challenges, including overfitting to training data, limited generalization to long-tail and cross-domain scenarios, and the lack of interpretability. For instance, a stop sign displayed on an advertisement screen may mislead autonomous vehicles into abnormal driving behaviors, highlighting a key limitation of purely data-driven approaches that focus on direct input–output mapping without knowledge induction and causal reasoning.

The emergence of generalized human-like intelligence in recent advances of Large Language Models (LLMs)~\cite{qwen3-llm, qwen2-llm, intern-llm} and Vision-Language Models (VLMs)~\cite{Qwen25-vl, qwen3-vl, llava, llava-ov, internvl, internvl-3.5, mobilevlm_v2} offers a promising bridge to leverage extensive knowledge for addressing the aforementioned challenges. Pioneering studies have explored replacing traditional end-to-end models with VLM-based architectures, taking advantage of their reasoning and linguistic capabilities to enable knowledge-driven autonomous driving~\cite{wisead, emma, drivemm, drivevlm, drivemlm, drivelm, lmdrive}. However, most existing approaches adopt a supervised fine-tuning (SFT) paradigm and rely heavily on fixed answer patterns, which limits their flexibility and generalization in linguistic representation and semantic diversity. Inspired by the success of reinforcement fine-tuning (RFT)~\cite{rft}, particularly the recent advances in DeepSeek-R1~\cite{deepseek-r1}, Group Relative Policy Optimization (GRPO)~\cite{GRPO} has been introduced in several early studies~\cite{autovla, recogdrive, alphadrive} to strengthen reasoning ability. Nevertheless, these works primarily focus on verifiable tasks such as meta-decision making~\cite{alphadrive} or trajectory planning~\cite{autovla, recogdrive}, where outputs are deterministic and easily quantifiable. In contrast, real-world autonomous driving involves numerous open-ended reasoning tasks. For example, in response to the question \textit{``Why are you decelerating?''}, multiple contextually reasonable answers may exist. This reveals an important and underexplored challenge: \textit{How can we effectively integrate open-ended driving knowledge with concrete driving behaviors to extend reasoning and driving performance beyond fixed ground-truth supervision?}

To tackle this challenge, we propose \OURS, an \textbf{OPEN}-ended \textbf{RE}asoning reinforced vision-language model (VLM) designed for autonomous driving (\textbf{AD}). OpenREAD handles a broad spectrum of tasks, ranging from open-ended reasoning such as perception and scene understanding to verifiable driving behaviors such as trajectory planning. This capability is achieved through a joint end-to-end RFT paradigm over both open-ended knowledge and planning datasets, enhancing generalization and reasoning capabilities while improving downstream planning performance. Inspired by the training protocol of DeepSeek-R1~\cite{deepseek-r1}, we begin with cold-start initialization. GPT-4~\cite{gpt-4} is employed to generate reasoning annotations on a subset of LingoQA~\cite{LingoQA}, a large-scale open-ended driving knowledge dataset covering scenario perception and action-related tasks. We further incorporate the OmniDrive dataset~\cite{omnidrive} by converting its annotations into reasoning-based textual trajectories. After equipping the model with foundational driving knowledge through supervised fine-tuning (SFT), we perform end-to-end RFT using GRPO~\cite{GRPO} to further enhance reasoning quality. For open-ended questions without explicit ground truth, Qwen3-8B serves as a critic to assess the generated responses with continuous rewards. In addition, rewards based on semantic similarity and average displacement error (ADE) are introduced to encourage the model to learn to reason and drive rather than merely imitating ground-truth answers or behaviors.

Extensive experiments demonstrate that joint end-to-end RFT substantially improves both upstream reasoning and downstream planning, enabling \OURS to achieve state-of-the-art performance on reasoning and planning benchmarks. Our main contributions are summarized as follows:
\begin{enumerate}
    \item We present a thinking-enabled autonomous driving model that employs RFT across both knowledge learning and trajectory planning, extending the end-to-end learning paradigm beyond traditional SFT.
    
    \item We introduce an LLM-as-critic mechanism that allows GRPO to be applied not only to verifiable tasks but also to open-ended knowledge learning, making end-to-end RL feasible for autonomous driving.
    
    \item We construct a set of Chain-of-Thought (CoT) datasets designed to equip autonomous driving models with essential domain knowledge and explicit reasoning capability.

\end{enumerate}
\begin{figure*}[t]
\centering
\includegraphics[width=0.95\textwidth]{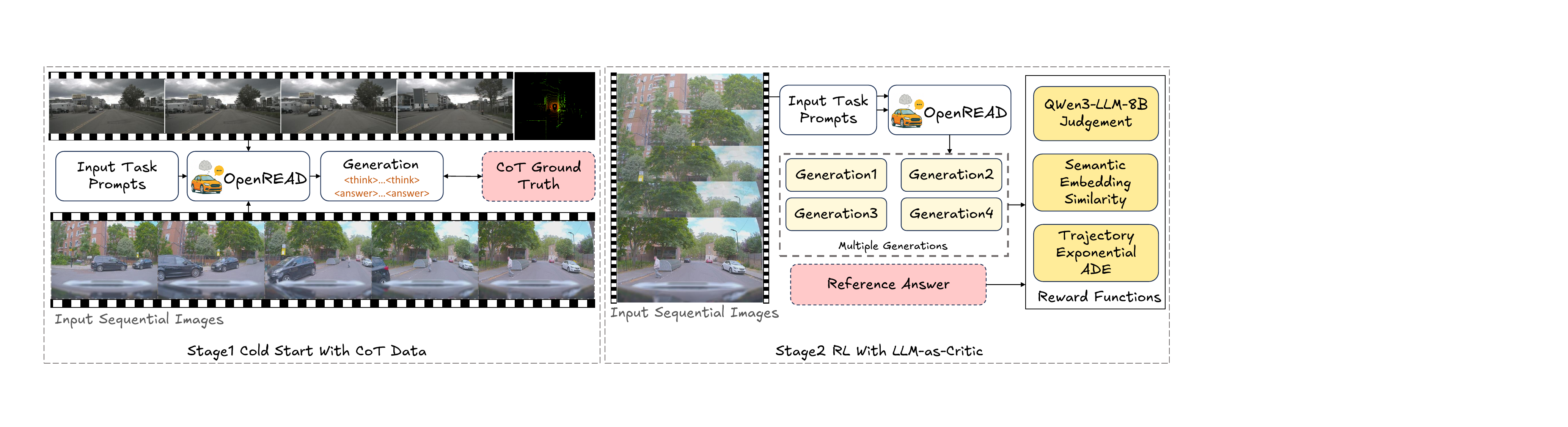}
\caption{The training pipeline of our \OURS. For the cold start stage, we utilize the CoT annotated data for SFT, followed by RFT with GRPO to further enhance the reasoning capabilities.}
\label{fig:framework}
\end{figure*}
\vspace{-1mm}

\section{Related Works}
\subsection{End-to-End Autonomous Driving}
End-to-end autonomous driving abandons the traditional pipeline of separate perception, prediction and planning modules in favor of a single learning-based framework that maps raw sensor inputs directly to motion plans or control actions \cite{li2023towards, wormann2022knowledge}. Such approaches have recently gained traction thanks to large-scale driving datasets, improved simulation tools and deep learning methods for joint feature optimization \cite{uniad, jia2023think, chen2024end, vadv2}. DriveTransformer \cite{drivetransformer} unifies perception, prediction, and planning within a single Transformer architecture, enabling direct cross-task and sensor-to-task attention for efficient and scalable end-to-end driving. ParaDrive \cite{paradrive} emphasizes the interdependence of sub-tasks and proposes a parallelized structure to improve coordination between prediction and planning. DiffusionDrive \cite{diffusiondrive} introduces a generative diffusion model that produces diverse and continuous trajectory candidates, enhancing motion diversity under uncertain conditions. Despite these advances, existing end-to-end methods still fall short in capturing high-level scene semantics and complex agent interactions, which limits their ability to generalize to diverse, long-tail, or previously unseen driving scenarios \cite{koma}.

\subsection{LLMs and VLMs for Autonomous Driving}
The recent rise of LLMs and VLMs has sparked growing interest in bringing semantic reasoning and general world knowledge into autonomous driving. Early explorations such as ADAPT \cite{adapt}, DriveGPT4\cite{drivegpt4} and LMDrive \cite{lmdrive} adapt VLMs for tasks like scene captioning, reasoning, or language-guided control, demonstrating the feasibility of integrating linguistic priors into autonomous driving. Subsequent systems like DriveVLM \cite{drivevlm}, DriveMLM \cite{drivemlm}, RAG-Driver \cite{RAG}, and ELM \cite{elm} expanded this direction by incorporating multi-modal perception or retrieval-augmented knowledge, but most models either rely solely on supervised fine-tuning or provide only high-level decisions without producing actionable trajectories.

More recent work has begun leveraging VLMs for deeper semantic reasoning that directly supports motion-related predictions \cite{dilu, safedrive}. SimLingo \cite{simlingo} aligns natural-language instructions with driving actions by combining VLM-based language understanding, instruction-conditioned policy learning, and closed-loop trajectory generation from camera-only input. ORION \cite{orion} integrates long-context visual encoding with an LLM-based reasoning module to enrich scenario interpretation before generating trajectories. OpenDriveVLA \cite{opendrivevla} proposes a language-driven supervision and evaluation pipeline, demonstrating the feasibility of text-conditioned motion assessment. AutoVLA \cite{autovla} introduces autoregressive action modeling within a VLM backbone and shows that structured reasoning signals can benefit downstream planning under reinforcement fine-tuning. ReCogDrive \cite{recogdrive} adopts a cognition-to-planning pipeline where VQA-aligned language features condition a diffusion-based motion predictor. On the data side, OmniDrive \cite{omnidrive} provides large-scale video QA and counterfactual reasoning annotations built on nuScenes \cite{nuscenes}, supplying strong supervision for causal and semantic understanding.


\subsection{Reinforcement Fine-Tuning for LLMs and VLMs}
RFT has recently emerged as an effective paradigm for improving the reasoning capability of LLMs and VLMs. Methods such as Vision Think \cite{visionthink} and Visual RFT \cite{visualrft} extend RFT to visual or multimodal settings, allowing models to refine intermediate reasoning steps under verifiable rewards. Vision-R1 \cite{visionr1} further adapts the DeepSeek-R1 paradigm \cite{deepseek-r1} to vision-based tasks, showing that stepwise reward-driven optimization can significantly enhance visual logical reasoning. Reason-RFT \cite{reasonrft} generalizes this idea to broader multimodal problems by coupling structured reasoning traces with reward-based updates, improving consistency and factuality.
These works collectively demonstrate that reward-driven reasoning optimization can substantially improve the semantic understanding and decision quality of large models. In autonomous driving, where both open-ended scene reasoning and motion-related predictions require multi-step inference, such reinforcement tuning strategies offer a promising mechanism for jointly enhancing high-level reasoning and downstream decision-making.

\section{Methodolgy}
\subsection{Overview of \OURS}
Our \OURS is built on top of the Qwen3-VL-8B model~\cite{qwen3-vl}. The overall framework and training pipeline is illustrated in Fig.~\ref{fig:framework}. Given a sequence of images along with the task-specific textual prompts as input, our \OURS is capable of generating the corresponding answers with an explicit thinking process. Specifically, we use five sequential images of the front view for open-ended driving knowledge learning and four sequential images along with a Lidar Bird-Eye's View image of the current timestep for the trajectory planning task, as inspired by~\cite{transfuser, drivemm, drivemlm} to integrate the surrounding 3D information. The fine-tuning process involves two stages: cold start with the CoT annotated data in a SFT, followed by the RFT with GRPO algorithm. We will give a detailed description of each stage in the following sections.

\subsection{Cold Start}
\subsubsection{CoT Data Generation}\label{sec:cot_data}
To endow the model with fundamental reasoning and thinking capabilities, we start with CoT annotations for cold start. Specifically, we adopt the LingoQA dataset~\cite{LingoQA} as our primary source of driving-related knowledge, which provides high-quality question–answer pairs covering both scenario understanding and action analysis tasks. We manually annotate 100 samples with concise rationales that focus key objects within each scenario, avoiding redundant reasoning contents. These annotated examples are subsequently provided to GPT-4.1 to summarize the target annotation prompt, as illustrated in the left panel of Fig.~\ref{fig:qwen3_prompt}. 

\begin{figure}[ht]
\centering
\includegraphics[width=0.45\textwidth]{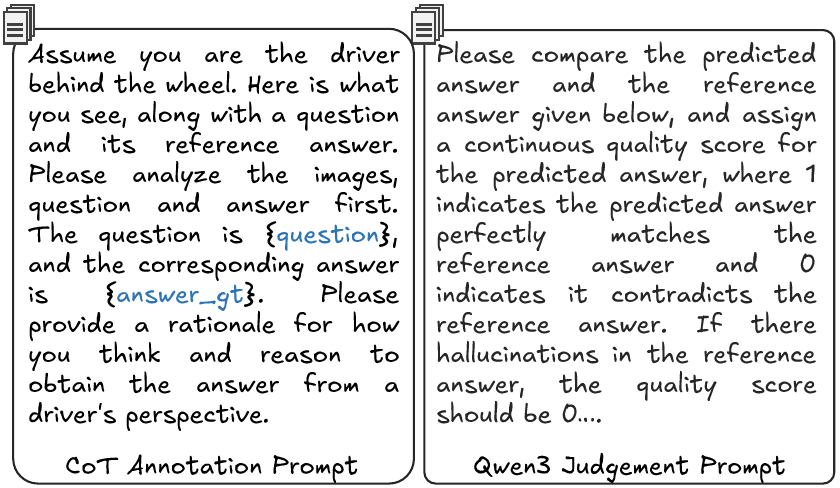}
\caption{\textbf{An overview of the prompt templates} used for CoT annotation generations (Left) and Qwen3-LLM open-ended driving knowledge evaluation (Right).}
\label{fig:qwen3_prompt}
\end{figure}
\vspace{-1mm}

\begin{figure*}[ht]
\centering
\includegraphics[width=0.95\textwidth]{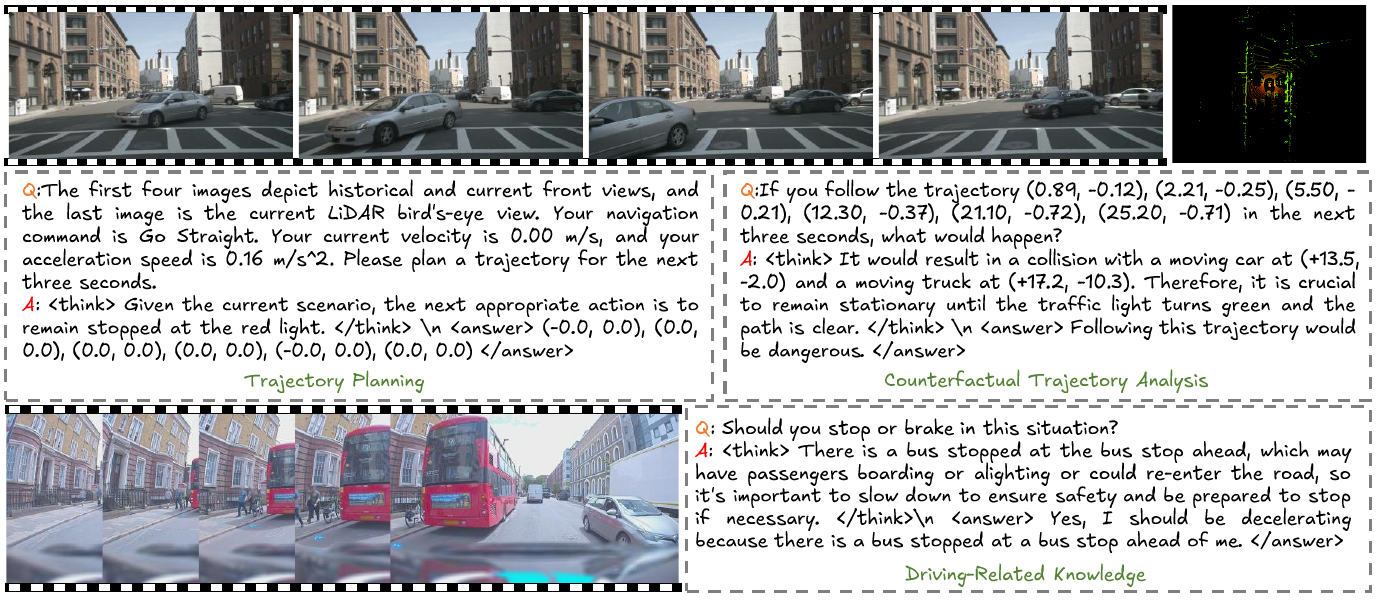}
\caption{An overview of our CoT annotations for driving-related knowledge and trajectory planning.}
\label{fig:data_anno}
\end{figure*}

We then leverage the derived annotation prompt and provide GPT-4.1 with the corresponding images, questions, and reference answers to generate target rationales from a driver's perspective. The placeholder of \textit{\{question\}} and \textit{\{answer\_gt\}} will be updated accordingly during the generation process. After manually post-processing to refine or discard inappropriate outputs, we obtain a total of 7K high-quality CoT annotations for driving knowledge on the LingoQA dataset, as in Fig.~\ref{fig:data_anno}.

Moreover, we incorporate the OmniDrive dataset, a well-annotated CoT dataset built on the NuScenes dataset. We separate the original mixture of reasoning and answer contents, and reformat the data into our target structure by enclosing the reasoning within \textit{\textless{}think\textgreater{}...\textless{}/think\textgreater{}} 
and the ground-truth answers within \textit{\textless{}answer\textgreater{} ...\textless{}/answer\textgreater{}}. The modified OmniDrive dataset integrates two tasks: trajectory planning and counterfactual trajectory analysis. Representative examples are illustrated in Fig.~\ref{fig:data_anno}. For the trajectory planning task, the reasoning component describes the high-level meta action and its underlying rationale, while the answer provides a future trajectory of six points over the next three seconds at 2 Hz. For the counterfactual trajectory analysis task, a pseudo trajectory is given, and the objective is to determine whether this trajectory is reasonable. This task helps the model to have a more comprehensive understanding of the planning and improves the spatial perception capabilities of the surroundings. The total amount of CoT annotations for the trajectory planning and counterfactual trajectory analysis tasks is 24K and 7K, respectively.

\subsubsection{SFT with CoT Data}
Given a sequence of $N$ images $X_{\text{v}} \in \matR^{N \times H \times W \times 3}$, a set of visual tokens $H_{\text{v}} \in \matR^{N \times N_\text{v} \times D_\text{v}}$ is obtained from the vision encoder and the projector, where $N_\text{v}$ and $D_\text{v}$ denote the visual token number and the feature dimension for each image, respectively. Conditioned on the textual instruction $X_{\text{t}}$ and its corresponding textual embeddings $H_{\text{t}}$, the LLM computes the probability of the target answer $Y_{\text{a}}$ in autoregression:

\begin{equation}
p(Y_{\text{a}}|H_{\text{v}}, H_{\text{t}}) = \prod_{i=1}^{L}{p(y_i|H_\text{v}, H_\text{t}, y<i)},
\end{equation}

\noindent where $L$ is the length of the target answer. The cross-entropy function is then applied to compute the loss between the prediction and ground-truth.

\subsection{RFT with LLM-as-Critic}
\subsubsection{Preliminary on GRPO Algorithm}
To further enhance the reasoning capability of \OURS, we introduce a second RFT stage based on the GRPO algorithm~\cite{GRPO}. In contrast to traditional reinforcement learning methods such as Proximal Policy Optimization (PPO)~\cite{PPO}, which rely on a critic network for policy evaluation, GRPO simplifies the reward mechanism by directly comparing groups of candidate responses. Given a query $q$ that integrates the visual input and task-specific instruction, GRPO samples a set of outputs ${o_1, o_2, \ldots, o_G}$ from the previous policy $\pi_{\theta_{\text{old}}}$ and computes the normalized group-relative advantage $A_i$:

\begin{equation}
A_i = \frac{r_i - \text{mean}(\{r_j\}_{j=1}^G)}{\text{std}(\{r_j\}_{j=1}^G)},
\end{equation}
where $r_i$ is the reward value for the $i$-th sample. The current policy $\pi_{\theta}$ is then optimized by maximizing the objective:

\begin{equation}
\label{grpo}
\mathcal{J}_{\text{GRPO}}(\theta) = \mathbb{E}_{q, \{o_i\} \sim \pi_{\theta_{\text{old}}}} \left[
\frac{1}{G} \sum_{i=1}^{G} \left(\mathcal{J}^{R}_i
- \beta \mathbb{D}_{\text{KL}}(\pi_\theta \| \pi_{\text{ref}})
\right)
\right],
\end{equation}
\begin{equation}
\mathcal{J}_i^{R} = 
\min \left(w_i A_i,\ 
\text{clip} \left(w_i, 1 - \epsilon, 1 + \epsilon \right) A_i
\right),
\end{equation}

\noindent where $w_i=\frac{\pi_\theta(o_i|q)}{\pi_{\theta_{\text{old}}}(o_i|q)}$, and $\epsilon$ and $\beta$ are hyper-parameters.

\subsubsection{Open-ended Knowledge Rewards}\label{sec:know_reward}
A key component of the RFT stage lies in the design of the reward function, which guides the model to efficiently explore and optimize its policy. To address the chellenge of open-ended knowledge rewards assignment discussed before, we employ Qwen3-LLM~\cite{qwen3-llm} as an automatic evaluator that functions as a classifier to assign reward scores. The key components of the user prompt are illustrated in the right panel of Fig.~\ref{fig:qwen3_prompt}. The \textit{predicted}\_ \textit{answer} and \textit{reference\_ answer} are updated accordingly during the training process. For each predicted response and its reference answer, Qwen3-LLM outputs a continuous value between 0 and 1 representing the probability that the predicted answer semantically matches the reference. This value is then discretized based on a predefined threshold, which is 0.5 in this work. The Qwen3-LLM reward $R_\text{qwen3}$ for the $i$-th predicted response $r_{q_i}$ is:

\begin{equation}
r_{{q_i}} =
\begin{cases}
1, & \text{if } p_i > 0.5, \\
0, & \text{else },
\end{cases}
\end{equation}

\noindent where $p_i$ denotes the probability assigned by Qwen3-LLM to the $i$-th predicted response. 

To further encourage the model to omit the redundant reasoning and produce concise outputs, we introduce an additional semantic-level reward, $R_{semantic}$. This reward is computed by extracting the sentence embeddings of both the predicted response and the reference answer by~\cite{minicpm}\footnote{We use a finetuned version of \url{https://huggingface.co/microsoft/MiniLM-L12-H384-uncased}.}, followed by calculating their cosine similarity, which ranges from 0 to 1. A higher similarity score indicates that the predicted response exhibits a closer alignment in expression, including sentence length, semantic meaning, and overall phrasing.

The overall reward for open-ended knowledge learning $R_\text{know}$ is a weighted combination of the Qwen3-LLM reward $R_\text{qwen3}$ and the semantic embedding similarity $R_\text{semantic}$:

\begin{equation}
R_{\text{know}} = w_{\text{q}} R_{\text{qwen3}} + w_{\text{s}}R_{\text{semantic}},
\end{equation}
where $w_{\text{q}}$ and $w_{\text{s}}$ are weighting hyperparameters set to 0.9 and 0.1, respectively.

\subsubsection{Trajectory Planning Reward}
Given the $i$-th sample of the planned trajectory and the ground truth over the next 3 seconds at 2Hz (resulting in 6 points in total), we compute the average $L_2$ distance for each second, denoted as $\{{d_i}_1, {d_i}_2, {d_i}_3\}$. The trajectory reward $R_\text{traj}$ for the current sample ${r_\text{traj}}_i$ is then calculated as:

\begin{equation}
r_{t_i} = 5^{-{d_i}_1*T_1}+e^{-{d_i}_2*T_2}+e^{-{d_i}_3*T_3},
\end{equation}

\noindent where $T_n$ is the temperature scalor for the $n$-th second. In our work, the temperatures are set to $0.7, 0.7, 0.3$ for corresponding 1s, 2s, and 3s.

\subsubsection{Joint Training}
During the RFT stage, the trajectory planning task is jointly trained with the learning of driving-related knowledge. The default format reward $R_\text{format}$ is also adopted to constrain the reasoning components enclosed in \textit{\textless{}think\textgreater{} ...\textless{}/think\textgreater{}} and the final response enclosed in \textit{\textless{}answer\textgreater{} ...\textless{}/answer\textgreater{}}. The overall reward is a combination of $R_\text{know}$ and $R_\text{traj}$ for the corresponding responses within a batch: 
\begin{equation}
R = R_\text{know} + R_\text{traj} + R_\text{format}.
\end{equation}

\section{Experiment}\label{sec:exp}
\begin{table*}[t]
\centering
\caption{
Effectiveness of extending driving-related knowledge via RFT for both trajectory planning and knowledge learning. Jointly learning trajectory planning and driving-related knowledge with enhanced reasoning capabilities through RFT substantially reduces the L2 distance error and improves knowledge acquisition, outperforming the SFT notably.
}
\resizebox{0.95\textwidth}{!}{
\begin{tabular}{lcccccccccccc}
\toprule
\multirow{3}{*}{\textbf{Data}} & 
\multirow{3}{*}{\textbf{Epoch}}& 
\multirow{3}{*}{\textbf{Type}} & 
\multicolumn{8}{c}{\textbf{Traj Eval}} & \multicolumn{2}{c}{\textbf{Know Eval}} \\
& & & \multicolumn{4}{c}{\textbf{L2 (m)} $\downarrow$} & \multicolumn{4}{c}{\textbf{Collision (\%)} $\downarrow$} & \textbf{Coun. Traj.}$\uparrow$ & \textbf{LingoQA}$\uparrow$ \\ \cline{4-11}
 & & & \textbf{1s} & \textbf{2s} & \textbf{3s} & \textbf{Avg.} & \textbf{1s} & \textbf{2s} & \textbf{3s} & \textbf{Avg.} & \textbf{GPT-Score} & \textbf{LingoJudge}  \\
\midrule
\multirow{3}{*}{Traj.} & 1 & SFT & 0.22 & 0.48 & 0.85 & 0.52 & 0.06 & 0.14 & 0.50 & 0.23 & - & - \\
& 1 & Cold-Start & 0.22 & 0.48 & 0.84 & 0.51 & 0.09 & 0.13 & 0.43 & 0.22 & - & - \\
& 1 & RFT & 0.23 & 0.48 & 0.83 & 0.51 & 0.05 & 0.18 & 0.49 & 0.24 & - & - \\ \midrule
\multirow{3}{*}{Traj.+Count.Traj.} & 1 & SFT & 0.20 & 0.43 & 0.73 & 0.45 & 0.05 & 0.10 & 0.37 & 0.17 & 62.2\% & -\\
& 1 & Cold-Start & 0.21 & 0.44 & 0.74 & 0.46 & \textbf{0.03} & 0.11 & 0.36 & 0.17 & 67.0\% & -\\
& 1 & RFT & 0.20 & 0.42 & 0.68 & 0.43 & \textbf{0.03} & 0.07 & 0.30 & 0.13 & 67.4\% & - \\ \midrule

\multirow{4}{*}{\begin{tabular}{@{}c@{}}
Traj.+Count.Traj. \\
+LingoQA
\end{tabular}} & 1 & SFT & 0.20 & 0.44 & 0.76 & 0.47 & 0.06 & 0.11 & 0.37 & 0.18 & 65.6\% & 66.4\% \\
& 2 & SFT & 0.20 & 0.42 & 0.71 & 0.44 & \textbf{0.03} & \textbf{0.07} & 0.29 & 0.13 & 68.0\% & 67.4\% \\
& 1 & Cold-Start & 0.19 & 0.41 & 0.71 & 0.44 & 0.04 & 0.10 & 0.30 & 0.15 & 68.2\% & 65.4\% \\
& 1 & RFT & \textbf{0.18} & \textbf{0.37} & \textbf{0.63} & \textbf{0.40} & 0.04 & \textbf{0.07} & \textbf{0.20} &\textbf{ 0.11} & \textbf{68.8\%} & \textbf{68.2}\% \\ \bottomrule
\end{tabular}
}
\label{tab:nuscenes_ablation}
\end{table*}

\subsection{Implementation Details}
We leverage the training framework of Model-Swift~\cite{model_swift}, an official framework provided by the ModelScope community for fine-tuning and deploying large language models and multi-modal large models. Our \OURS is implemented on top of the Qwen3-VLM-8B~\cite{qwen3-vl} dense model. For the cold-start stage, we conduct supervised fine-tuning with a total batch size of 32 on 4 NVIDIA-H100 GPUs. The learning rate is set to $1\times10^{-4}$ with a warm-up ratio of 0.05, followed by a cosine decay schedule. It takes one epoch of fine-tuning on the CoT-annotated datasets introduced in Sec.~\ref{sec:cot_data}, which include driving-related knowledge and trajectory planning on the LingoQA,~\cite{LingoQA}, and OmniDrive~\cite{omnidrive} datasets. The training data amounts for the cold start stage are 7K, 9K, and 24K for LingoQA, counterfactual trajectory analysis, and trajectory planning, respectively.

The RFT stage is initialized from the cold start checkpoint. To maintain a balanced learning ratio between trajectory planning and driving-related reasoning, we randomly sample 6K trajectory planning examples and 5K mixed samples from LingoQA and counterfactual trajectory data. For each training step, the number of response generations is set to 4. The learning rate is set to $1\times10^{-4}$ with a total batch size of 16. We employ LoRA~\cite{lora} and utilize the AdamW~\cite{adamw} optimizer during both the SFT and RFT stages.

\subsection{Evaluation Metric}
For driving-related knowledge evaluation, we report the official Lingo-Judge metric for the LingoQA dataset which uses a pretrained text transformer to assess the semantic correctness of the predicted answer with respect to the question and reference response. We also report a GPT-score for evaluating the counterfactural trajectory analysis, which computes the average accuracy by utilizing the GPT-4.1 to evaluate whether the predicted response matches the reference. In addition, we follow previous works \citep{elm, LingoQA, llava} and report three established metrics of CIDEr \citep{CIDEr}, BLEU \citep{bleu}, and Meteor~\cite{meteor} which collectively measure the similarity between generated explanations and reference descriptions, covering content relevance, n-gram consistency, and semantic matching. For the trajectory planning task on the NuScenes dataset, we report the ST-P3~\cite{stp3} based metric in terms of the L2 distance error and collision rate across 1s, 2s, and 3s.

\subsection{RFT for Knowledge-Enhanced Planning}
We conduct thorough ablation experiments to assess the effectiveness of RFT and the necessity of extending driving-related knowledge. The results are presented in Tab.~\ref{tab:nuscenes_ablation}. When training on the trajectory-planning-only task, we use the full training set during the SFT and cold-start stages for one epoch. The RFT stage involves a set of 6K randomly sampled data. Due to the absence of driving-related knowledge data and relatively weak capabilities obtained at the cold-start stage, the subsequent RFT stage fails to deliver notable gains. In fact, the average collision rate even degrades at the 2-second and 3-second prediction. A similar degradation is also observed in AutoVLA~\cite{autovla}. These ablation studies indicate that performing RFT solely on trajectory planning provides only limited benefits, highlighting the necessity of extending driving-related knowledge.

To help the model achieve a more comprehensive understanding of the planning task, we further introduce the counterfactual trajectory analysis data. At this stage, we use the full set of trajectory planning data together with 9K counterfactual trajectory samples during the SFT and cold-start stages. During the RFT stage, we sample 6K trajectory-planning examples and 5K counterfactual trajectory analysis samples to achieve a balanced learning of planning ability and driving-related knowledge. The L2 distance and collision rate exhibit a significant improvement across the SFT, cold start, and RFT stages, strongly demonstrating the necessity of extending driving-related knowledge. It is worth noting that, when initialized from a well-established cold-start stage, the RFT training mechanism begins to exhibit a stronger potential at this stage. It outperforms the SFT by reducing the L2 distance from 0.45m to 0.43m and collision rate from 0.17\% to 0.13\%, respectively, demonstrating the effectiveness of knowledge extension via RFT in enhancing planning quality.

\begin{table*}[t]
\centering
\caption{
Comparison of the Open-loop planning in nuScene. The ST-P3 evaluation protocol is used by default.
$\dagger$: The ego status and planning trajectory are both processed by LLM in textual modality. 
$\ddagger$: Use a customized evaluation protocol instead of ST-P3 metrics.
}
\resizebox{0.99\textwidth}{!}{
\begin{tabular}{lccccccccccc}
\toprule
\multirow{2}{*}{\textbf{Method}} & \multirow{2}{*}{\textbf{VLM-Based}} & \multicolumn{2}{c}{\textbf{Ego Status}} & \multicolumn{4}{c}{\textbf{L2 (m)} $\downarrow$} & \multicolumn{4}{c}{\textbf{Collision (\%)} $\downarrow$} \\
\cline{3-12}
 & & \textbf{BEV} & \textbf{Planner} & \textbf{1s} & \textbf{2s} & \textbf{3s} & \textbf{Avg.} & \textbf{1s} & \textbf{2s} & \textbf{3s} & \textbf{Avg.} \\
\midrule
VAD~\cite{vad} & - & \checkmark & \checkmark & 0.17 & 0.34 & 0.60 & 0.37 & 0.07 & 0.10 & 0.24 & 0.14 \\
UniAD~\cite{uniad} & - & \checkmark & \checkmark & 0.44 & 0.67 & 0.96 & 0.69 & 0.04 & 0.08 & 0.23 & 0.12 \\
Ego-MLP~\cite{egomlp} & - & \checkmark & \checkmark & 0.46 & 0.76 & 1.12 & 0.78 & 0.21 & 0.35 & 0.58 & 0.38 \\ 
OpenDrive-VLA$^{\dagger}$~\cite{opendrivevla} & \checkmark & \checkmark & - & 0.15 & 0.31 & 0.55 & 0.33 & \textbf{0.01} & 0.08 & 0.21 & \textbf{0.10} \\
ORION\cite{orion}$^\ddagger$ & \checkmark & \checkmark & \checkmark & 0.17 & 0.31 & 0.55 & 0.34 & 0.05 & 0.25 & 0.80 & 0.37 \\ \midrule
ST-P3~\cite{stp3} & - & - & - & 1.33 & 2.11 & 2.90 & 2.11 & 0.23 & 0.62 & 1.27 & 0.71 \\
RoboTron-Drive$^{\dagger, \ddagger}$~\cite{drivemm} & \checkmark & - & - & 0.14 & 0.30 & 0.57 & 0.33 & 0.03 & 0.12 & 0.63 & 0.26 \\
EMMA$^{\dagger}$~\cite{emma} & \checkmark & - & - & \textbf{0.14} & \textbf{0.29} & \textbf{0.54} & \textbf{0.32} & - & - & - & - \\
OpenEMMA$^{\dagger}$~\cite{emma} & \checkmark & - & - & 1.45 & 3.21 & 3.76 & 2.81 & - & - & - & - \\
DriveVLM$^{\dagger}$~\cite{drivevlm} & \checkmark & - & - &  0.18 & 0.34 & 0.68 & 0.40 & 0.10 & 0.22 & 0.45 & 0.27 \\
GPT-Drive${^\dagger}$~\cite{gpt-driver} & \checkmark & - & - & 0.20 & 0.40 & 0.70 & 0.44 & 0.04 & 0.12 & 0.36 & 0.17 \\
OmniDrive~\cite{omnidrive}$^\ddagger$ & \checkmark & - & - & 0.40 & 0.80 & 1.32 & 0.84 & 0.04 & 0.46 & 2.32 & 0.94 \\
AutoVLA$^{\dagger}$~\cite{autovla} & \checkmark & - & - & 0.25 & 0.46 & 0.73 & 0.48 & 0.07 & \textbf{0.07} & 0.26 & 0.13 \\
\midrule
\OURS(Ours)$^{\dagger}$ & \checkmark & - & - & 0.18 & 0.37 & 0.63 & 0.40 & 0.04 & \textbf{0.07} & \textbf{0.20} & 0.11 \\ \bottomrule
\end{tabular}
}
\label{tab:openloop_planning}
\end{table*}

The introduction of the additional LingoQA dataset further extends the diversity of the driving knowledge, covering scene understanding and action analysis across a wide range of scenarios. The comprehensive questions about the object recognition, action justification, driving attention, \etc, help to improve the model with enhanced general perception and reasoning capabilities. The SFT training stage is performed on additional 10k LingoQA samples, while the CoT training stage uses 7K LingoQA with CoT annotations instead. Equipped with the reasoning capabilities, our \OURS efficiently obtains a notable improvement during the cold-start stage, which is further improved during the RFT stage by jointly learning the driving knowledge and planning through GRPO, outperforming the SFT with two epochs of training by an obvious margin. Besides the trajectory planning performance, extending the driving-related knowledge with enhanced reasoning capabilities also improves the acquisition of the knowledge, which is reflected by the improved GPT-score for the counterfactual trajectory analysis at the cold start stage. For the LingoQA dataset, the Lingo-Judge accuracy is also increased via RFT, increasing from 65.4\% to 68.2\%. 

The overall experiments on the NuScenes trajectory evaluation and knowledge evaluation highlight that empowering the model with accumulated driving-related knowledge and enhanced reasoning capability through GRPO-based RFT is key to achieving consistent improvements in planning performance and knowledge acquisition. This strongly validates the effectiveness of employing Qwen3-LLM as the judgment mechanism in our framework.

\subsection{State-of-the-Art Benchmark}
As shown in Tab.~\ref{tab:openloop_planning}, we conduct a comparison with other SOTA approaches. Due to the leverage of extensive internal training data, EMMA\cite{emma} achieves the lowest L2 distance error, even outperforming the traditional end-to-end models by a notable margin. Compared with other VLM-based models trained on a comparable data scale, our \OURS obtains competitive performance and yields the lowest collision rate in the final two seconds. It is worth noting that the comparison with AutoVLA~\cite{autovla} which is also trained via RFT but exclusively on the trajectory planning data, our method achieves an obvious improvement, further highlighting the necessity of joint learning with driving-related knowledge. For driving-related knowledge evaluation, we benchmark \OURS against both generalist and specialist models. As reported in Tab.~\ref{tab:know_eval}, only Qwen2.5-VL~\cite{Qwen25-vl} is able to generate concise responses, whereas others tend to produce overly verbose outputs, leading to degraded evaluation metrics. Compared with specialist models trained on the LingoQA dataset, \OURS achieves the highest Lingo-Judge accuracy, demonstrating the effectiveness of reinforcing reasoning capabilities through RFT.

We omit comparison results for counterfactual trajectory analysis because generalist models are unable to correctly interpret coordinate-based inputs, while specialist models were not trained on this task, resulting in poor performance across all baselines.

\begin{table}[t]
\centering
\renewcommand{\arraystretch}{1.1}
\setlength{\tabcolsep}{1pt}
\caption{Comparison with generalist and specialist VLMs on the LingoQA. Our \OURS achieves the highest Lingo-Judge accuracy. Generalist models tend to produce redundant responses, which results in significantly degraded performance.}
\resizebox{0.45\textwidth}{!}{\begin{tabular}{lcccc}
\toprule
\textbf{Method} & \textbf{Lingo-Judge} & \textbf{CIDEr} & \textbf{BLEU} & \textbf{Meteor} \\
\midrule
LLava-OV-7B~\cite{llava-ov} & 54.2\% & 0.39 & 0.03 & 0.08 \\
Qwen2.5-VL-7B~\cite{Qwen25-vl} & 62.6\% & 0.54 & 0.07 & 0.11 \\
Qwen3-VL-4B~\cite{qwen3-vl} & 37.0\% & 0.00 & 0.00 & 0.04 \\
Qwen3-VL-8B~\cite{qwen3-vl} & 51.4\% & 0.00 & 0.01 & 0.11 \\
InternVL3-5-8B~\cite{internvl-3.5} & 48.2\% & 0.14 & 0.02 & 0.12 \\ \midrule
OpenEMMA~\cite{emma} & 29.0\% & 0.07 & 0.02 & 0.11 \\
WiseAD~\cite{wisead} & 60.4\% & 0.68 & \textbf{0.20} & 0.19 \\
RoboTron-Drive~\cite{drivemm} & 59.2\% & 0.65 & 0.14 & 0.19 \\
ReCogDrive~\cite{recogdrive} & 67.8\% & 0.69 & 0.13 & \textbf{0.21} \\ \midrule
\textbf{\OURS(ours)} & \textbf{68.2\%} & \textbf{0.74} & 0.15 & 0.20 \\
\bottomrule
\end{tabular}}
\label{tab:know_eval}
\end{table}

\vspace{-1mm}

\subsection{Ablation Studies}
\subsubsection{Reward for Driving Knowledge Learning}
To validate the effectiveness of our proposed reward design for driving-related knowledge learning, we conduct ablation experiments on a fixed set of 10K samples from the LingoQA dataset. Our 7K CoT annotated data are built upon this 10K samples. Our 7K CoT-annotated data are constructed from this subset. We first investigate RFT initialized from the vanilla Qwen3-VLM, as reported in the first four rows of Tab.\ref{tab:coldstart_results}. Leveraging Qwen3-LLM as the only reward signal yields only marginal improvement, increasing the Lingo-Judge accuracy from 51.4\% to 52.0\%. This limited gain is largely due to the absence of constraints on response style: the generated outputs remain verbose and contain redundant descriptive content, similar to the vanilla Qwen3-VLM. Such redundancy is directly reflected in the degraded CIDEr, BLEU, and Meteor scores. Introducing the semantic embedding similarity reward can effectively alleviate the redundancy issue. However, embedding-based similarity primarily captures global sentence structure and often fails to discriminate key contradictions between reference answers and model predictions. For instance, the two sentences \textit{“I should decelerate”} and \textit{“I should accelerate”} yield a similarity score of 0.6. This ambiguity in reward assignment can degrade the reinforcement learning process, as evidenced by higher CIDEr, BLEU, and Meteor scores but a substantially lower Lingo-Judge accuracy. The weighted combination of these two rewards, as described in Sec.~\ref{sec:know_reward}, leads to improved accuracy while preserving the concise and straightforward expression style of the reference answers. This balance is reflected by consistent improvements across all evaluation metrics. Nevertheless, due to the lack of explicit guidance on the thinking and reasoning process, the overall performance still falls short of expectations. 

Following DeepSeek-R1~\cite{deepseek-r1}, we employ a cold start stage with explicit thinking guidance. Introducing the 7K CoT annotations with explicit reasoning steps effectively equips the model with fundamental reasoning capabilities, improving the Lingo-Judge accuracy from 51.4\% to 65.0\% while maintaining satisfactory expression quality. This highlights the critical role of the cold start stage. Subsequently, continuing RFT on the remaining 3K samples from a well-initialized checkpoint further facilitates the exploration process: using Qwen3-LLM as the judgment signal yields the best Lingo-Judge performance, and incorporating the weighted embedding similarity additionally helps constrain the overall expression style to better align with the reference answers, which is finally adopted in our work.
\begin{table}[t]
\centering
\renewcommand{\arraystretch}{1.1}
\setlength{\tabcolsep}{1pt}
\caption{Ablation on the reward design and cold start for knowledge learning through GRPO.}
\resizebox{0.45\textwidth}{!}{
\begin{tabular}{lcccccc}
\toprule
\textbf{Method} & \textbf{Type} & \textbf{Reward} & \textbf{Lingo-Judge}$\uparrow$ & \textbf{CIDEr}$\uparrow$ &  \textbf{BLEU}$\uparrow$ & \textbf{Meteor}$\uparrow$ \\
\midrule
\multirow{4}{*}{
\begin{tabular}{@{}c@{}}
w/o \\
Cold Start
\end{tabular}
} & Base & - & 51.4\% & 0.00 & 0.01 & 0.11 \\ \cline{2-7}
&\multirow{3}{*}{RL}
& Qwen3 & 52.0\% & 0.37 & 0.03 & 0.10 \\
& & Semantic & 52.4\% & 0.65 & 0.13 & 0.19 \\
& & Full & 59.2\% & 0.67 & 0.13 & 0.20  \\ \midrule
\multirow{4}{*}{
\begin{tabular}{@{}c@{}}
Cold Start \\
(7K)
\end{tabular}
} & Base & - & 65.0\% & 0.71 & 0.15 & 0.20 \\ \cline{2-7}
 & \multirow{4}{*}{RL} & Qwen3 & \textbf{66.4\%} & 0.72 & 0.14 & 0.20 \\
 & & Semantic & 63.2\% & 0.66 & 0.15 & 0.20 \\
 & & Full & 66.0\% & \textbf{0.75} & \textbf{0.15} & \textbf{0.21} \\ \bottomrule
\end{tabular}
}

\label{tab:coldstart_results}
\vspace{-1mm}
\end{table}

\begin{table}[t]
    \centering
    \setlength\tabcolsep{1pt}
    \renewcommand\arraystretch{1.3}
    \caption{Ablation on the data scale for knowledge learning on the LingoQA. Increasing the data amount introduces a constant improvement across different stages.}
    \resizebox{0.45\textwidth}{!}{
    \begin{tabular}{lccccc} \toprule
    \textbf{Method} & \textbf{Data} & \textbf{Lingo-Judge}$\uparrow$ & \textbf{CIDEr} $\uparrow$ & \textbf{BLEU} $\uparrow$ & \textbf{Meteor}$\uparrow$  \\ \midrule
    \multirow{2}{*}{SFT} & 10k & 65.6\% & 0.765 & 0.149 & 0.206  \\
    & 50k & 68.0\% & \textbf{0.773} & 0.147 & \textbf{0.208} \\ \midrule
    \multirow{2}{*}{Cold Start} & 3k & 63.0\% & 0.664 & 0.138 & 0.201 \\
    & 7k & 65.0\% & 0.705 & 0.153 & 0.203 \\ \hline
    \multirow{3}{*}{GRPO} & 3k & 66.0\% & 0.753 & 0.154 & 0.205 \\
    & 5k & 67.2\% & 0.739 & 0.165 & 0.201 \\ 
    & 10k & \textbf{68.4\%} & 0.756 & \textbf{0.168} & 0.202 \\ \bottomrule
    \end{tabular}
    }

\label{tab:cold_start_data_amount}
\end{table}

\subsubsection{Data Scale for Knowledge Learning}
As shown in Tab.~\ref{tab:cold_start_data_amount}, we explore the effectiveness of the training data amount for SFT, cold-start, and RFT. Clearly, increasing the amount of training data helps to obtain a constant gain across different training stages. Compared with SFT that requires 50K training data to improve the Lingo-Judge from 65.6\% to 68.0\%, our \OURS takes advantage of the GRPO algorithm to explore multiple responses within a rollout, increasing the Lingo-Judge accuracy from 65.0\% to 68.4\% with only another 10K data after the cold-start stage.

\section{Conclusion and Limitations}
In this work, we present \OURS, an open-ended reasoning reinforced VLM designed for joint knowledge learning and trajectory planning through RFT paradigm. By incorporating Qwen3-LLM as the critic in RL loop, our approach enables effective reinforcement of open-ended knowledge through GRPO, leading to SOTA performance in both trajectory planning and knowledge evaluation. Due to the computational cost associated with multiple generations from \OURS and Qwen3-LLM, the current scale of datasets used for RFT training remains limited. We will further scale up the knowledge data in future research.

{
    \small
    \bibliographystyle{ieeenat_fullname}
    \bibliography{main}
}

\clearpage
\setcounter{page}{1}
\maketitlesupplementary

\section{Visualization on the Trajectory Planning}
\begin{figure}[h]
\centering
\includegraphics[width=0.38\textwidth]{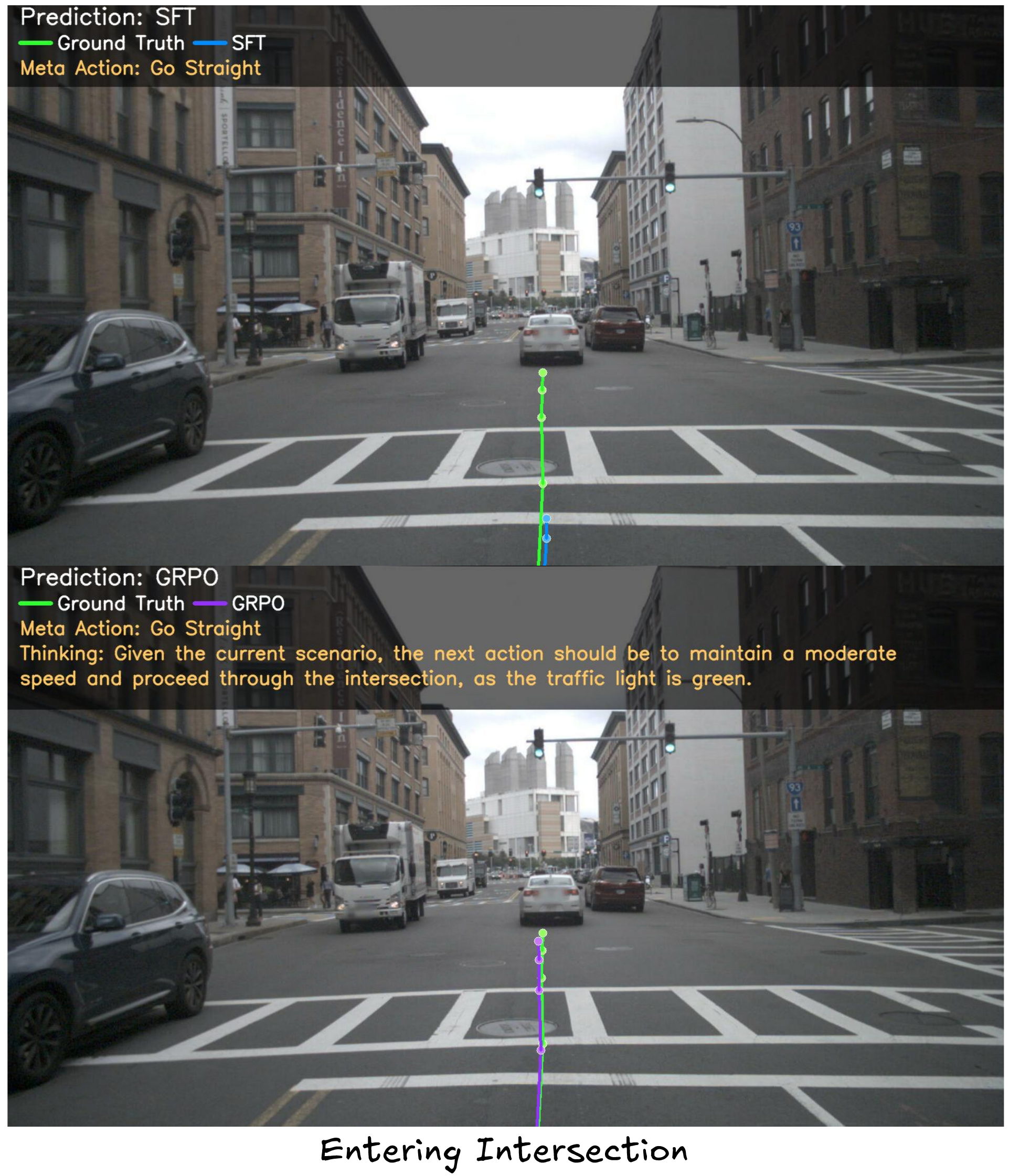}
\caption{OpenREAD performs a less-conservative planning when entering the intersection.}
\label{fig:qwen3_prompt}
\vspace{-1mm}
\end{figure}

\begin{figure}[h]
\centering
\includegraphics[width=0.38\textwidth]{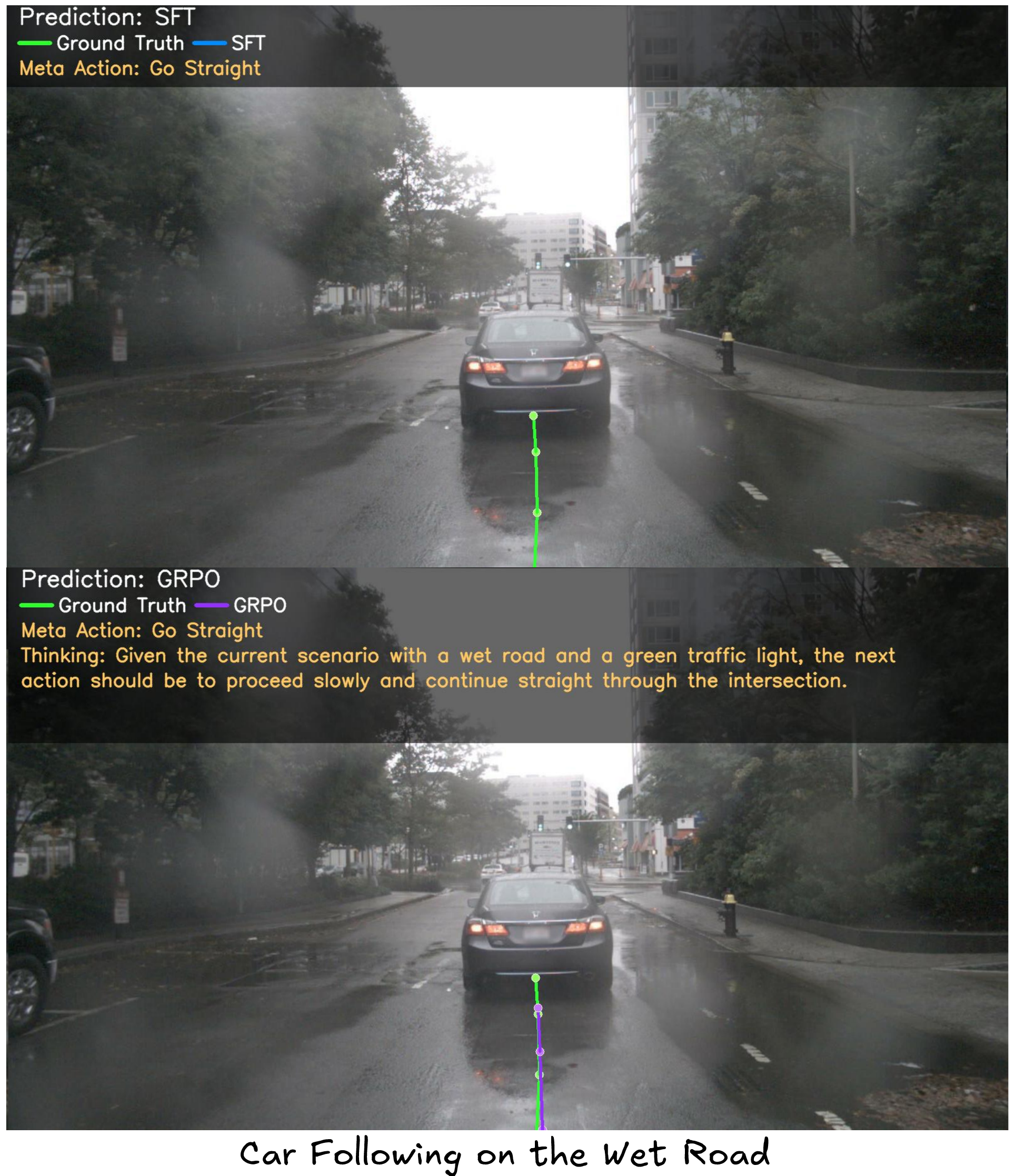}
\caption{OpenREAD performs a safe and consistent following on the wet road.}
\label{fig:qwen3_prompt}
\end{figure}

\begin{figure}[h]
\centering
\includegraphics[width=0.38\textwidth]{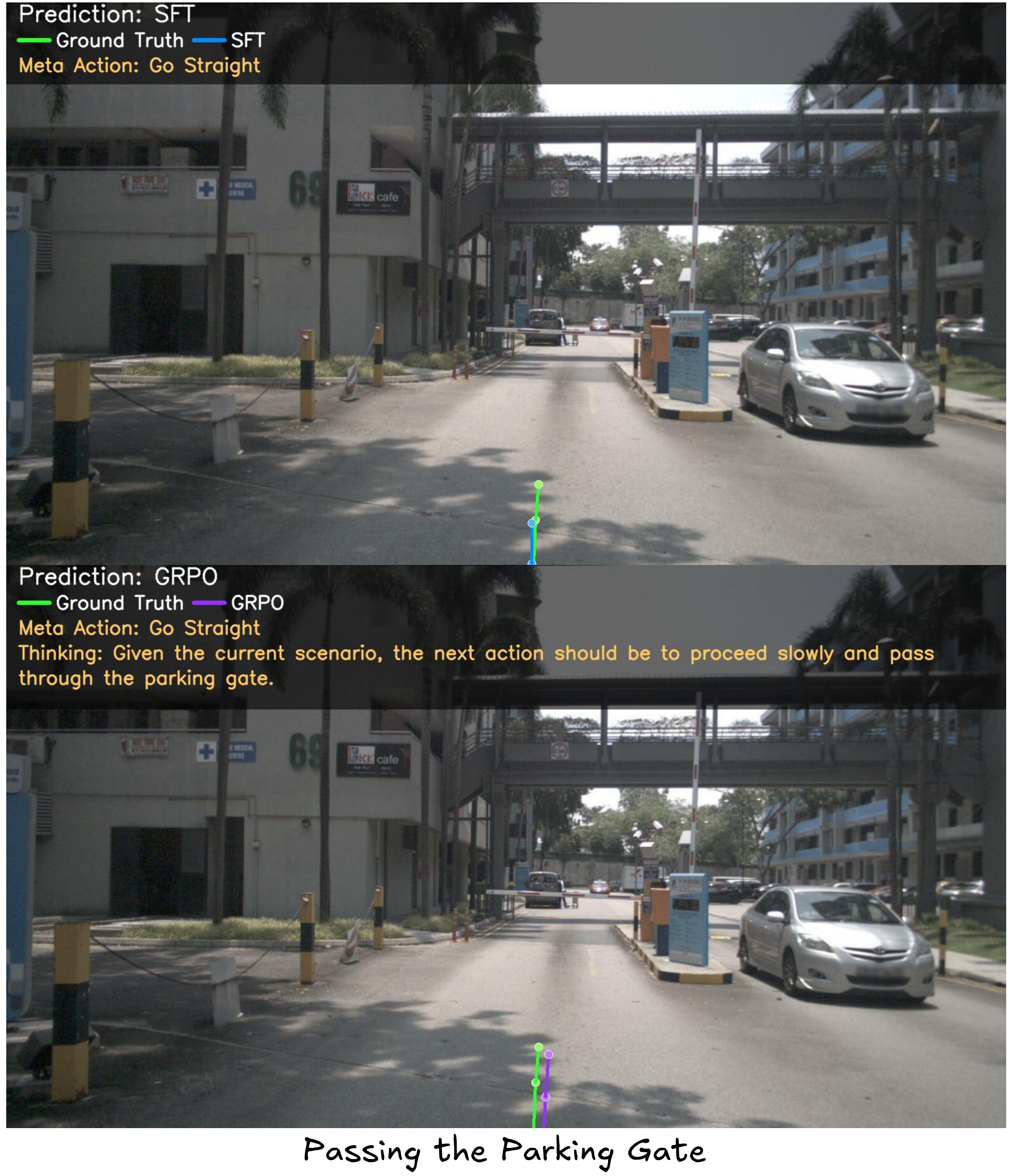}
\caption{OpenREAD is capable of recognizing the parking gate with cautious planning.}
\label{fig:qwen3_prompt}
\end{figure}

\begin{figure}[h]
\centering
\includegraphics[width=0.38\textwidth]{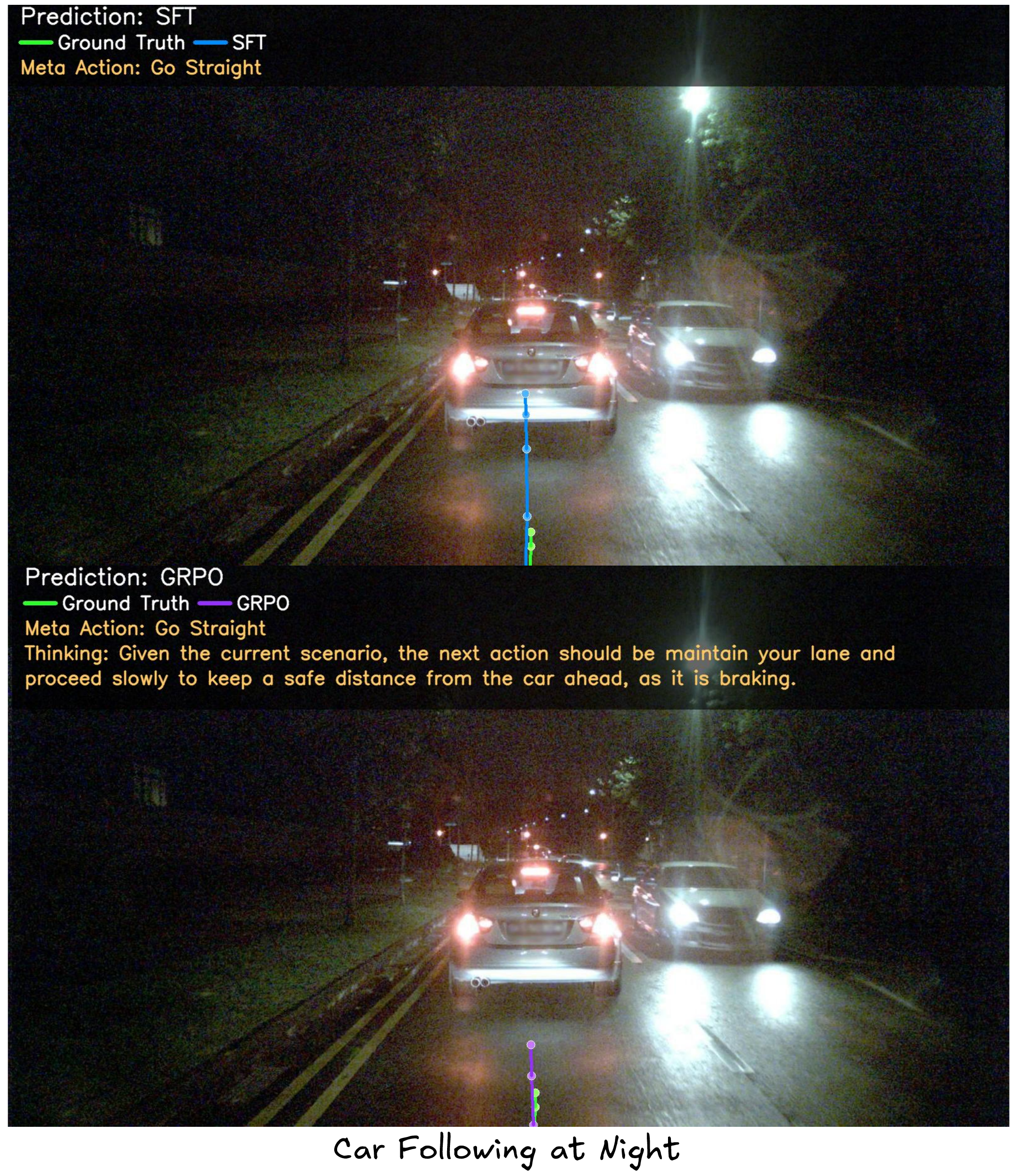}
\caption{OpenREAD plans a slow trajectory to avoid collision with the front car at night.}
\label{fig:qwen3_prompt}
\end{figure}

\section{Visualization on the Open-Ended Knowledge}

\begin{figure*}[h]
\centering
\includegraphics[width=0.95\textwidth]{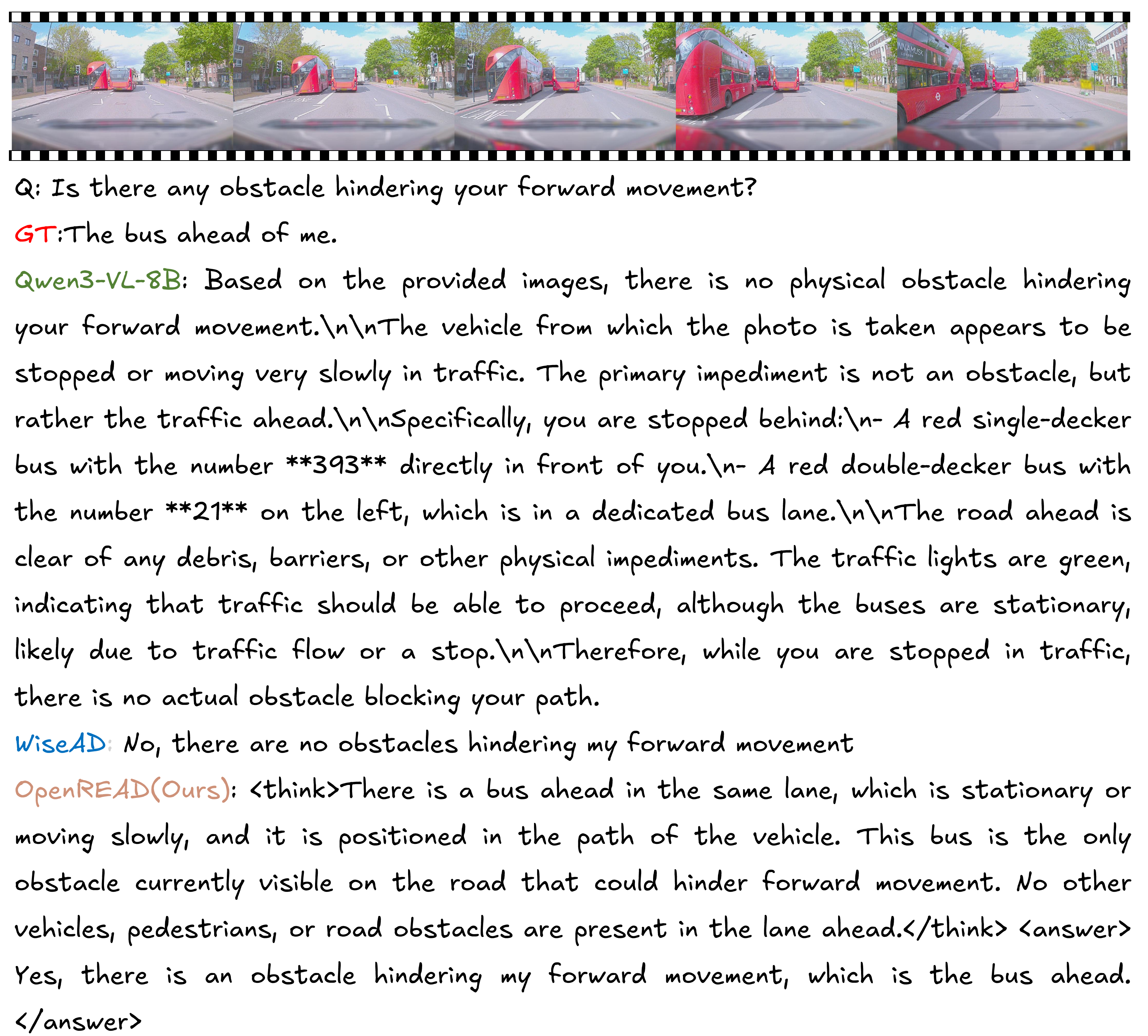}
\end{figure*}

\begin{figure*}[h]
\centering
\includegraphics[width=0.95\textwidth]{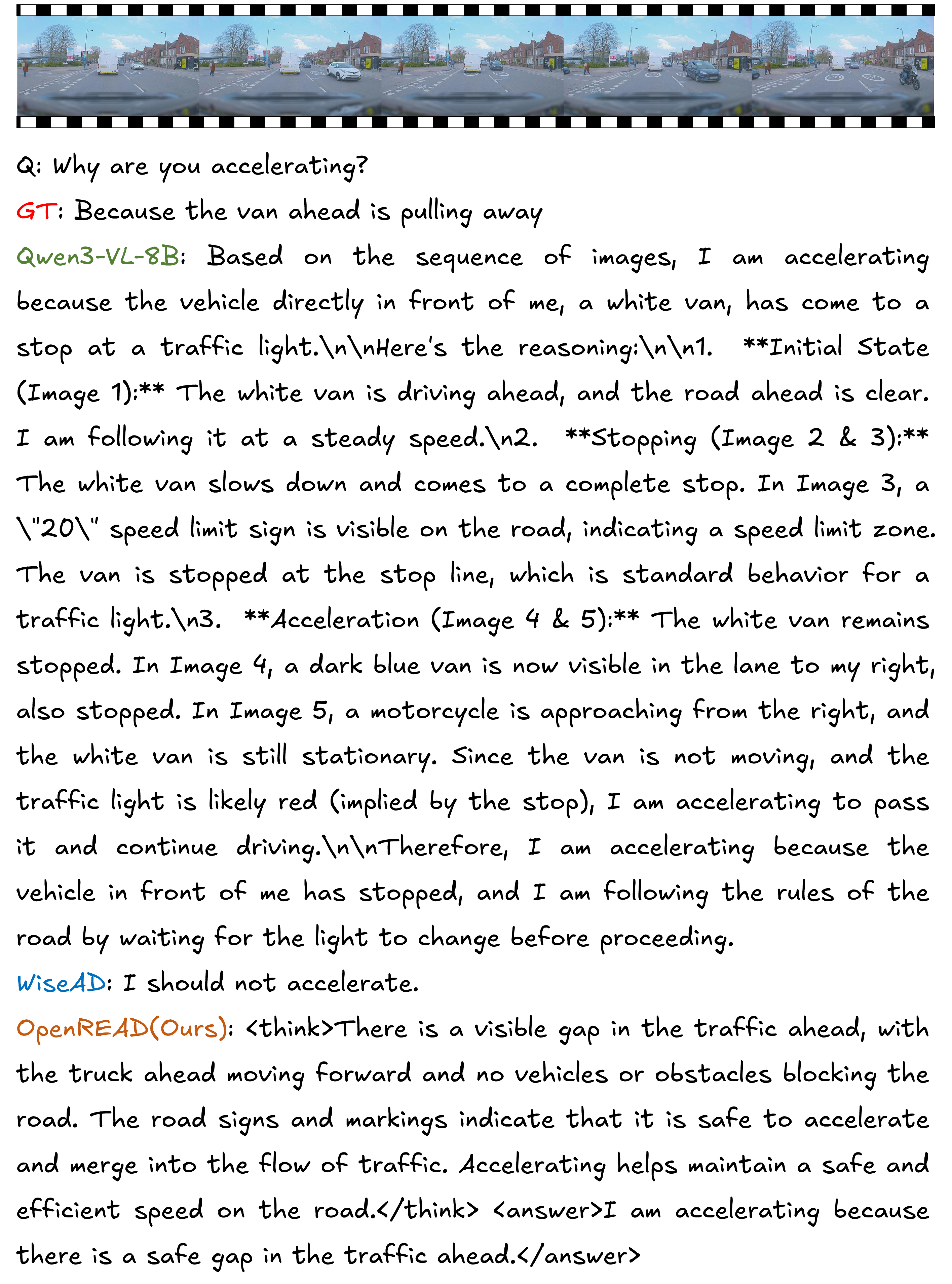}
\end{figure*}

\begin{figure*}[h]
\centering
\includegraphics[width=0.95\textwidth]{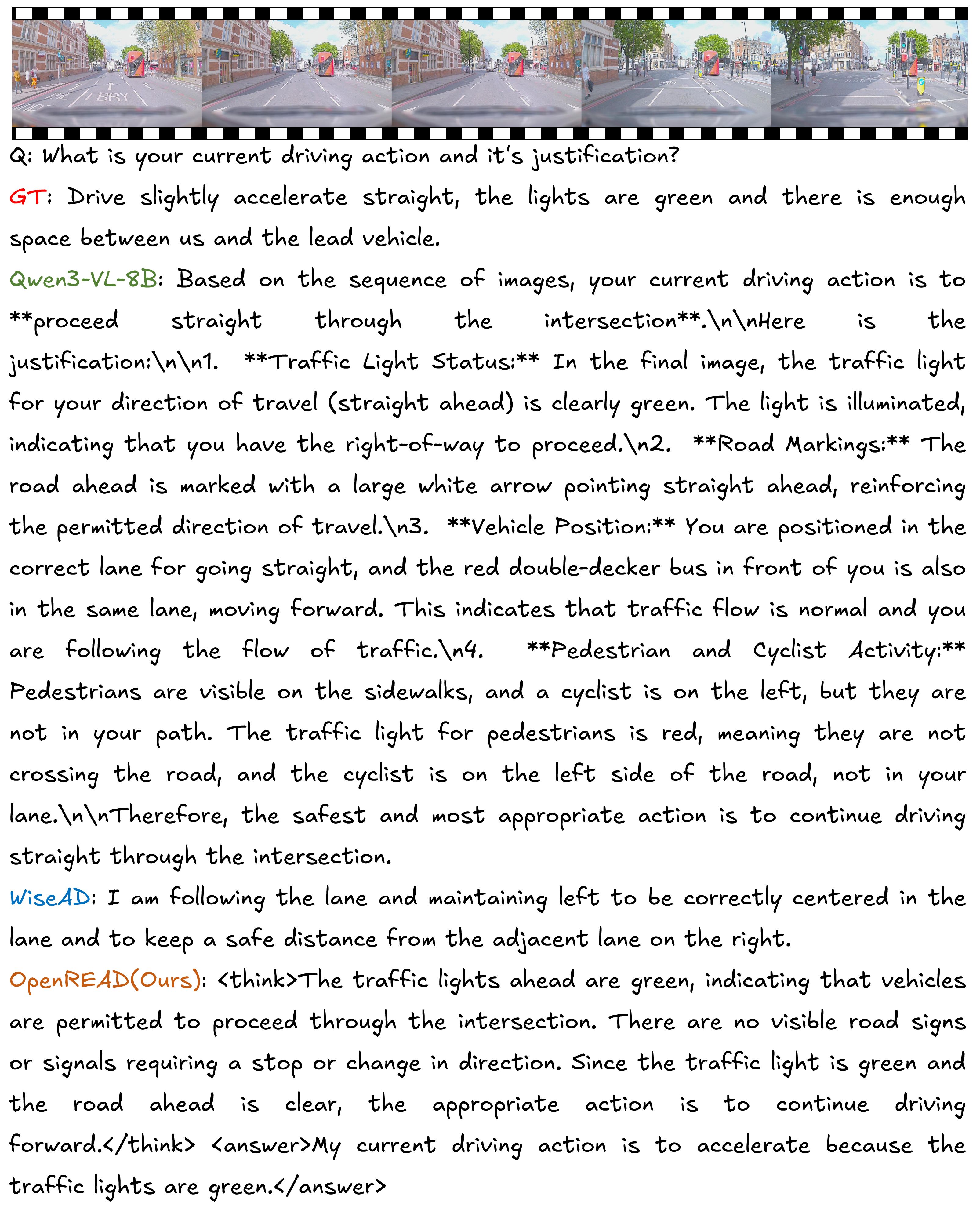}
\end{figure*}




\end{document}